\documentclass[runningheads, hidelinks]{llncs}
\pdfoutput=1 
 
\usepackage[T1]{fontenc}

\usepackage{graphicx}

\usepackage{hyperref}
\usepackage{blindtext}
\usepackage{makecell}
\usepackage{multirow}
\usepackage{array}
\newcolumntype{P}[1]{>{\centering\arraybackslash}p{#1}}
\newcolumntype{M}[1]{>{\centering\arraybackslash}m{#1}}
\usepackage{booktabs}
\usepackage{pifont}
\usepackage{bbding}
\usepackage{color}
\usepackage{xcolor, colortbl}

\definecolor{Gray}{gray}{0.9}
\usepackage{placeins}
\hypersetup{pdfauthor={Philip M\"uller and Felix Meissen and Johannes Brandt and Georgios Kaissis and Daniel Rueckert},pdftitle={Anatomy-Driven Pathology Detection on Chest X-rays}}

\begin{document}

\title{Anatomy-Driven Pathology Detection \\ on Chest X-rays}

\author{Philip Müller\inst{1} \and Felix Meissen\inst{1} \and Johannes Brandt\inst{1} \and Georgios Kaissis\inst{1,2} \and Daniel Rueckert\inst{1,3}}

\authorrunning{Philip Müller et al.}

\institute{Institute of AI in Medicine, Technical University of Munich, Munich, Germany \\
\email{philip.j.mueller@tum.de}
\and Helmholtz Zentrum Munich, Munich, Germany \and Department of Computing, Imperial College London, London, UK}

\maketitle 

\begin{abstract}
Pathology detection and delineation enables the automatic interpretation of medical scans such as chest X-rays while providing a high level of explainability to support radiologists in making informed decisions.
However, annotating pathology bounding boxes is a time-consuming task such that large public datasets for this purpose are scarce. Current approaches thus use weakly supervised object detection to learn the (rough) localization of pathologies from image-level annotations, which is however limited in performance due to the lack of bounding box supervision. 
We therefore propose  anatomy-driven pathology detection (ADPD), which uses easy-to-annotate bounding boxes of anatomical regions as proxies for pathologies. 
We study two training approaches: supervised training using anatomy-level pathology labels and multiple instance learning (MIL) with image-level pathology labels. Our results show that our anatomy-level training approach outperforms weakly supervised methods and fully supervised detection with limited training samples, and our MIL approach is competitive with both baseline approaches, therefore demonstrating the potential of our approach. 
\end{abstract}

\section{Introduction}
Chest radiographs (chest X-rays) represent the most widely utilized type of medical imaging examination globally and hold immense significance in the detection of prevalent thoracic diseases, including pneumonia and lung cancer, making them a crucial tool in clinical care \cite{raoof2012interpretation,johnson2019mimicphysio}. 
Pathology detection and localization -- for brevity we will use the term \emph{pathology detection} throughout this work -- enables the automatic interpretation of medical scans such as chest X-rays by predicting bounding boxes for detected pathologies. Unlike classification, which only predicts the presence of pathologies, it provides a high level of explainability supporting radiologists in making informed decisions.

However, while image classification labels can be automatically extracted from electronic health records or radiology reports \cite{chexpert,cxr8}, this is typically not possible for bounding boxes, thus limiting the availability of large datasets for pathology detection. Additionally, manually annotating pathology bounding boxes is a time-consuming task, further exacerbating the issue. 
The resulting scarcity of large, publicly available datasets with pathology bounding boxes limits the use of supervised methods for pathology detection, such that current approaches typically follow weakly supervised object detection approaches, where only classification labels are required for training. 
However, as these methods are not guided by any form of bounding boxes, their performance is limited.

We, therefore, propose a novel approach towards pathology detection that uses anatomical region bounding boxes, solely defined on anatomical structures, as proxies for pathology bounding boxes. 
These region boxes are easier to annotate and generalize better than those of pathologies, such that huge labeled datasets are available \cite{wu2021chest}.
In summary:
\begin{itemize}
    \item We propose anatomy-driven pathology detection (ADPD), a pathology detection approach for chest X-rays, trained with pathology classification labels together with anatomical region bounding boxes as proxies for pathologies.
    \item We study two training approaches: using localized (anatomy-level) pathology labels for our model \emph{Loc-ADPD} and using image-level labels with multiple instance learning (MIL) for our model \emph{MIL-ADPD}.
    \item We train our models on the Chest ImaGenome \cite{wu2021chest} dataset and evaluate on NIH ChestX-ray 8 \cite{cxr8}, where we found that our Loc-ADPD model outperforms both, weakly supervised methods and fully supervised detection with a small training set, while our MIL-ADPD model is competitive with supervised detection and slightly outperforms weakly supervised approaches. 
\end{itemize}
\section{Related Work}
\paragraph{Weakly Supervised Pathology Detection}
Due to the scarcity of bounding box annotations, pathology detection on chest X-rays is often tackled using weakly supervised object detection with Class Activation Mapping (CAM) \cite{cam}, which only requires image-level classification labels.
After training a classification model with global average pooling (GAP), an activation heatmap is computed by classifying each individual patch (extracted before pooling) with the trained classifier, before thresholding this heatmap for predicting bounding boxes. 
Inspired by this approach, several methods have been developed for chest X-rays \cite{chexnet,cxr8,stl,wsup_thoracic}. While CheXNet \cite{chexnet} follows the original approach, the method provided with the NIH ChestX-ray 8 dataset \cite{cxr8} and the STL method \cite{stl} use Logsumexp (LSE) pooling \cite{lse}, while the MultiMap model \cite{wsup_thoracic} uses max-min pooling as first proposed for the WELDON \cite{weldon} method.
Unlike our method, none of these methods utilize anatomical regions as proxies for predicting pathology bounding boxes, therefore leading to inferior performance. 

\paragraph{Localized Pathology Classification}
Anatomy-level pathology labels have been utilized before to train localized pathology classifiers \cite{wu2021chest,anaxnet} or to improve weakly supervised pathology detection \cite{agxnet}.
Along with the Chest ImaGenome dataset \cite{wu2021chest} several localized pathology classification models have been proposed which use a Faster R-CNN \cite{faster_rcnn} to extract anatomical region features before predicting observed pathologies for each region using either a linear model or a GCN model based on pathology co-occurrences. 
This approach has been further extended to use GCNs on anatomical region relationships \cite{anaxnet}.
While utilizing
the same form of supervision as our method, these methods 
do not tackle pathology detection. 

In AGXNet \cite{agxnet}, anatomy-level pathology classification labels are used to train a weakly-supervised pathology detection model. Unlike our and the other described methods, it does however not use anatomical region bounding boxes. 

\begin{figure}[t!]
    \centering
    \includegraphics[width=.9\linewidth]{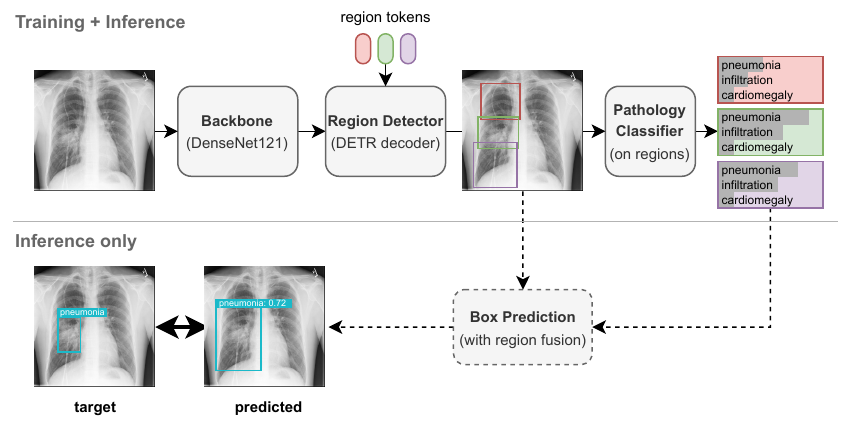}
    \caption{Overview of our method. Anatomical regions are first detected using a CNN backbone and a shallow detector. For each region, observed pathologies are predicted using a shared classifier.
    Bounding boxes for each pathology are then predicted by considering regions with positive predictions and fusing overlapping boxes.}
    \label{fig:overview}
\end{figure}

\section{Method}
\subsection{Model}
Fig.\ \ref{fig:overview} provides an overview of our method. 
Given a chest X-ray, we apply a DenseNet121 \cite{densenet} backbone and extract patch-wise features by using the feature map after the last convolutional layer (before GAP). We then apply a lightweight object detection model consisting of a single DETR \cite{DETR} decoder layer to detect anatomical regions. 
Following \cite{DETR}, we use learned query tokens attending to patch features in the decoder layer, where each token corresponds to one predicted bounding box. As no anatomical region can occur more than once in each chest X-ray, 
each query token is assigned to exactly one pre-defined anatomical region, such that the number of tokens equals the number of anatomical regions. This one-to-one assignment of tokens and regions allows us to remove the Hungarian matching used in \cite{DETR}. As described next, the resulting per-region features from the output of the decoder layer will be used for predictions on each region. 

For predicting whether the associated region is present, we use a binary classifier with a single linear layer, for bounding box prediction we use a three-layer MLP followed by sigmoid.
We consider the prediction of observed pathologies as a multi-label binary classification task and use a single linear layer (followed by sigmoid) to predict the probabilities of all pathologies.
Each of these predictors is applied independently to each region with their weights shared across regions. 

We experimented with more complex pathology predictors like an MLP or a transformer layer but did not observe any benefits. 
We also did not observe improvements when using several decoder layers and observed degrading performance when using ROI pooling to compute region features.

\subsection{Inference}
\begin{figure}[t!]
    \centering
    \includegraphics[width=.85\linewidth]{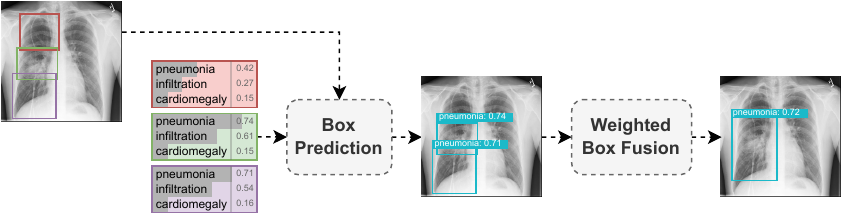}
    \caption{Inference. For each pathology, the regions with pathology probability above a threshold are predicted as bounding boxes, which are then fused if overlapping.}
    \label{fig:inference}
\end{figure}
As shown in Fig.\ \ref{fig:inference}, we predict pathology bounding boxes using the anatomical region boxes and per-region pathology probabilities in two steps.
In step (i), pathology probabilities are first thresholded and for each positive pathology (with probability larger than the threshold) the bounding box of the corresponding anatomical region is predicted as its pathology box, using the pathology probability as box score. This means, if a region contains several predicted pathologies, then all of its predicted pathologies share the same bounding box during step (i). 
In step (ii), weighted box fusion (WBF) \cite{wbf} merges 
bounding boxes of the same pathology with IoU-overlaps above 0.03 and computes weighted averages (using box scores as weights) of their box coordinates.
As many anatomical regions are at least partially overlapping, and we use a small IoU-overlap threshold, this allows the model to either pull the predicted boxes to relevant subparts of an anatomical region or to predict that pathologies stretch over several regions. 

\subsection{Training}
\begin{figure}[t!]
    \centering
    \includegraphics[width=.9\linewidth]{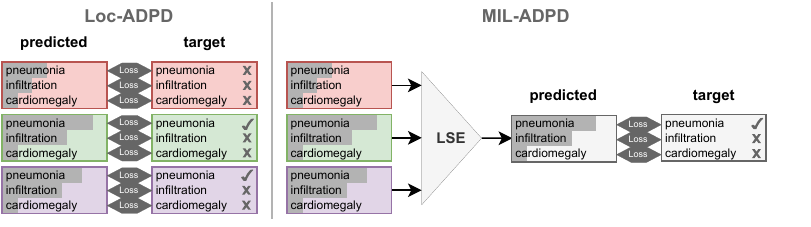}
    \caption{Training. \textbf{Loc-ADPD:} Pathology predictions of regions are directly trained using anatomy-level supervision. \textbf{MIL-ADPD:} Region predictions are first aggregated using LSE pooling and then trained using image-level supervision.}
    \label{fig:training}
\end{figure}
The anatomical region detector is trained using the DETR loss \cite{DETR} with fixed one-to-one matching (i.e.\ without Hungarian matching). 
For training the pathology classifier, we experiment with two different levels of supervision (Fig. \ref{fig:training}).

For our \emph{Loc-ADPD} model, we utilize anatomy-level pathology classification labels. 
Here, the target set of observed pathologies is available for each anatomical region individually such that the pathology observation prediction can directly be trained for each anatomical region. We apply the ASL \cite{asl} loss function independently on each region-pathology pair and average the results over all regions and pathologies. The decoder feature dimension is set to 512.

For our \emph{MIL-ADPD} model, we experiment with a weaker form of supervision, where pathology classification labels are only available on the per-image level.
We utilize multiple instance learning (MIL), where 
an image is considered a bag of individual instances (i.e.\ the anatomical regions), and only a single label (per pathology) is provided for the whole bag, which is positive if any of its instances is positive.
To train using MIL, we first aggregate the predicted pathology probabilities of each region over all detected regions in the image using LSE pooling \cite{lse}, acting as a smooth approximation of max pooling. The resulting per-image probability for each pathology is then trained using the ASL \cite{asl} loss. In this model, the decoder feature dimension is set to 256.

In both models, the ASL loss is weighted by a factor of 0.01 before adding it to the DETR loss.
We train using AdamW \cite{adamw} with a learning rate of 3e-5 (Loc-ADPD) or 1e-4 (MIL-ADPD) and weight decay 1e-5 (Loc-ADPD) or 1e-4 (MIL-ADPD) in batches of 128 samples with early stopping (with 20\,000 steps patience)
for roughly 7 hours on a single Nvidia RTX A6000.

\subsection{Dataset}
\paragraph{Training Dataset}
We train on the Chest ImaGenome dataset \cite{wu2021chest,wu2021chestphysio,PhysioNet}\footnote{\url{https://physionet.org/content/chest-imagenome/1.0.0} (PhysioNet Credentialed Health Data License 1.5.0)}, consisting of roughly 240\,000 frontal chest X-ray images with corresponding scene graphs automatically constructed from free-text radiology reports. It is derived from the MIMIC-CXR dataset \cite{johnson2019mimic,johnson2019mimicphysio}, which is based on imaging studies from 65\,079 patients performed at Beth Israel Deaconess Medical Center in Boston, US.
Amongst other information, each scene graph contains bounding boxes for 29 unique anatomical regions with annotated attributes, where we consider positive  \texttt{anatomical finding} and \texttt{disease} attributes as positive labels for pathologies, leading to binary anatomy-level annotations for 55 unique pathologies. We consider the image-level label for a pathology to be positive if any region is positively labeled with that pathology.

We use the provided jpg-images \cite{johnson2019mimicjpg}\footnote{\url{https://physionet.org/content/mimic-cxr-jpg/2.0.0/} (PhysioNet Credentialed Health Data License 1.5.0)} and follow the official MIMIC-CXR training split but only keep samples containing a scene graph with at least five valid region bounding boxes, resulting in a total of 234\,307 training samples. 
During training, we use random resized cropping with size 224x224, apply contrast and brightness jittering, random affine augmentations, and Gaussian blurring.

\paragraph{Evaluation Dataset and Class Mapping}
We evaluate our method on the NIH ChestXray-8 (CXR8) dataset \cite{cxr8}\footnote{\url{https://www.kaggle.com/datasets/nih-chest-xrays/data} (CC0: Public Domain)}, containing 108\,948 X-ray images from the National Institutes of Health Clinical Center in the US. 
We only consider the subset of 882 images 
with pathology bounding boxes, 
where we use 50\% for validation and keep the other 50\% as a held-out test set. 
All images are center-cropped and resized to $224 \times 224$.

The dataset contains bounding boxes for 8 unique pathologies. While partly overlapping with the training classes, a one-to-one correspondence is not possible for all classes. For some evaluation classes, we therefore use a many-to-one mapping where the class probability is computed as the mean over several training classes. We refer to Appendix \ref{app:mapping} for a detailed study on class mappings.

\section{Experiments and Results}
\subsection{Experimental Setup and Baselines}
We compare our method against several weakly supervised object detection methods (CheXNet \cite{chexnet}, STL \cite{stl}, GradCAM \cite{gradcam}, CXR \cite{cxr8}, WELDON \cite{weldon}, MultiMap Model \cite{wsup_thoracic}, LSE Model \cite{lse}), trained on the CXR8 training set 
using only image-level pathology labels.  
We also use AGXNet \cite{agxnet} for comparison, a weakly supervised method trained using anatomy-level pathology labels but without any bounding box supervision. It was trained on MIMIC-CXR (sharing the images with our method) with labels from RadGraph \cite{radgraph} and finetuned on the CXR8 training set with image-level labels. 
Additionally, we also compare with a Faster-RCNN \cite{faster_rcnn} trained on a small subset of roughly 500 samples from the CXR8 training set that have been annotated with pathology bounding boxes by two medical experts, including one board-certified radiologist. 

\begin{table}[ht!]
\centering
\caption{Results on the NIH ChestX-ray 8 dataset \cite{cxr8}. Our models Loc-ADPD and MIL-ADPD, trained using anatomy (An) bounding boxes, both outperform all weakly supervised methods trained with image-level pathology (Pa) and anatomy-level pathology (An-Pa) labels by a large margin. MIL-ADPD is competitive with the supervised baseline trained with pathology (Pa) bounding boxes, while Loc-ADPD outperforms it by a large margin.}
    \label{tab:results}
    \small
\setlength{\tabcolsep}{1.5pt}
\centering
\begin{tabular}{@{}lccccccccc@{}}
        \toprule
        \multirow{2}{*}{Method} &\multicolumn{2}{c}{Supervision} & \multicolumn{1}{c}{IoU@10-70} & \multicolumn{2}{c}{IoU@10} & \multicolumn{2}{c}{IoU@30} & \multicolumn{2}{c@{}}{IoU@50} \\
        \cmidrule(lr){2-3} \cmidrule(lr){4-4} \cmidrule(lr){5-6} \cmidrule(lr){7-8} \cmidrule(lr){9-10} & Box & Class & mAP & AP & loc-acc& AP & loc-acc & AP & loc-acc  \\
        \midrule
        MIL-ADPD (ours) & An & Pa & 7.84 & 14.01 & 0.68 & 8.85 & 0.65 & 7.03 & 0.65 \\ 
        \hspace{\parindent} w/o WBF &&& 5.42 & 11.05 & 0.67 & 7.97 & 0.65 & 3.44 & 0.64 \\
        Loc-ADPD (ours) & An & An-Pa & \textbf{10.89} & \textbf{19.99} & \textbf{0.85} & \textbf{12.43} & \textbf{0.84} & \textbf{8.72} & \textbf{0.83} \\ 
        \hspace{\parindent} w/o WBF &&& 8.88 & 17.02 & 0.84 & 9.65 & 0.83 & 7.36 & 0.83 \\
        \hspace{\parindent} w/ MIL &&& 10.29 & 19.16 & 0.84 & 10.95 & 0.83 & 8.00 & 0.82\\ 
        \midrule
        CheXNet \cite{chexnet} & - & Pa & 5.80 & 12.87 & 0.58 & 8.23 & 0.55 & 3.12 & 0.52 \\
        STL \cite{stl} & - & Pa  & 5.61 & 12.76 & 0.57 & 7.94 & 0.54 & 2.45 & 0.50 \\
        GradCAM \cite{gradcam} & - & Pa  & 4.43 & 12.53 & 0.58 & 6.67 & 0.54 & 0.13 & 0.51 \\ 
        CXR \cite{cxr8} & - & Pa  & 5.61 & 13.91 & 0.59 & 8.01 & 0.55 & 1.24 & 0.51 \\ 
        WELDON \cite{weldon} & - & Pa  & 4.76 & 14.57 & 0.61 & 6.18 & 0.56 & 0.34 & 0.51\\ 
        MultiMap \cite{wsup_thoracic} & - & Pa  & 4.91 & 12.36 & 0.61 & 7.13 & 0.57 & 1.35 & 0.53\\ 
        LSE Model \cite{lse} & - & Pa  & 3.77 & 14.49 & 0.61 & 2.62 & 0.56 & 0.42 & 0.54 \\ 
        \midrule
        AGXNet \cite{agxnet} & - & An-Pa & 5.30 & 11.39 & 0.59 & 6.58 & 0.56 & 4.14 & 0.54\\ 
        \midrule
        Faster R-CNN & Pa & - & 7.36 & 9.11 & 0.79 & 7.62 & 0.79 & 7.26 & 0.78 \\
        \bottomrule
    \end{tabular}
\end{table}

For all models, we only consider the predicted boxes with the highest box score per pathology, as the CXR8 dataset never contains more than one box per pathology.
We report the average precision (AP) at different IoU-thresholds and the mean AP (mAP) over thresholds $(0.1, 0.2, \dots, 0.7)$, commonly used thresholds on this dataset \cite{cxr8}.
Additionally, we report the localization accuracy (loc-acc) \cite{cxr8}, where we use a box score threshold of 0.7 for our method.

\subsection{Pathology Detection Results}
\paragraph{Comparison with Baselines}
Tab.\ \ref{tab:results} shows the results of our MIL-ADPD and Loc-ADPD models and all baselines on the CXR8 test set.
Compared to the best weakly supervised method with image-level supervision (CheXNet) our methods improve by large margins (MIL-ADPD by $\Delta +35.2 \%$, Loc-ADPD by $\Delta +87.8 \%$ in mAP). Improvements are especially high when considering larger IoU-thresholds and huge improvements are also achieved in loc-acc at all thresholds.
Both models also outperform AGXNet (which uses anatomy-level supervision) by large margins (MIL-ADPD by $\Delta +47.9 \%$ and Loc-ADPD by $\Delta +105.5 \%$ mAP), while improvements on larger thresholds are smaller here.
Even when compared to Faster R-CNN trained on a small set of fully supervised samples, MIL-ADPD is competitive ($\Delta +6.5 \%$), while Loc-ADPD improves by $\Delta +48.0 \%$. However, on larger thresholds (IoU@50) the supervised baseline slightly outperforms MIL-ADPD, while Loc-ADPD is still superior. This shows that using anatomical regions as proxies is an effective approach to tackle pathology detection. While using image-level annotations (MIL-ADPD) already gives promising results, the full potential is only achieved using anatomy-level supervision (Loc-ADPD).
Unlike Loc-ADPD and MIL-ADPD, all baselines were either trained or finetuned on the CXR8 dataset, showing that our method generalizes well to unseen datasets and that our class mapping is effective. 

For detailed results per pathology we refer to Appendix \ref{app:detailed_results}. We found that the improvements of MIL-ADPD are mainly due to improved performance on Cardiomegaly and Mass detection, while Loc-ADPD consistently outperforms all baselines on all classes except Nodule, often by a large margin.

\paragraph{Ablation Study}
In Tab.\ \ref{tab:results} we also show the results of different ablation studies. Without WBF, results degrade for both of our models, highlighting the importance of merging region boxes. 
Combining the training strategies of Loc-ADPD and MIL-ADPD does not lead to an improved performance. Different class mappings between training and evaluation set are studied in Appendix \ref{app:mapping}.

\paragraph{Qualitative Results}
\begin{figure}[t]
\small
\setlength{\tabcolsep}{0.5pt}
    \centering
    \begin{tabular}{cccc}
    \includegraphics[height=.24\textwidth]{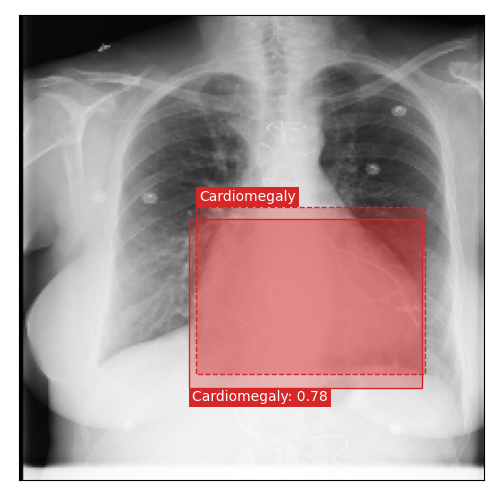} & 
    \includegraphics[height=.24\textwidth]{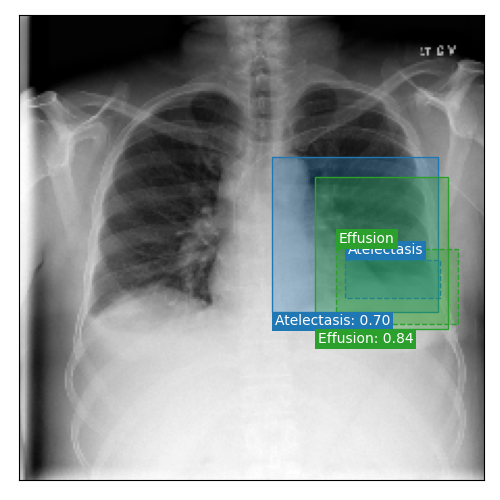} & \includegraphics[height=.24\textwidth]{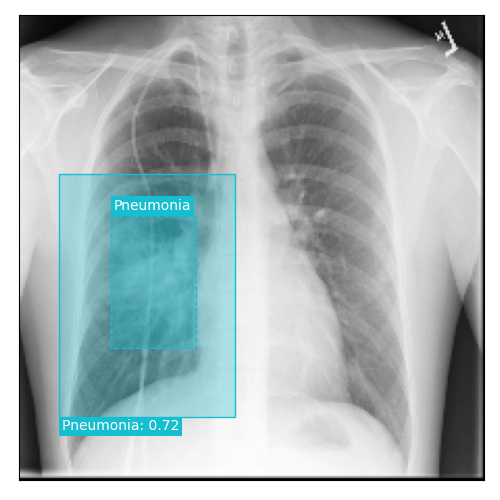}  &
    \includegraphics[height=.24\textwidth]{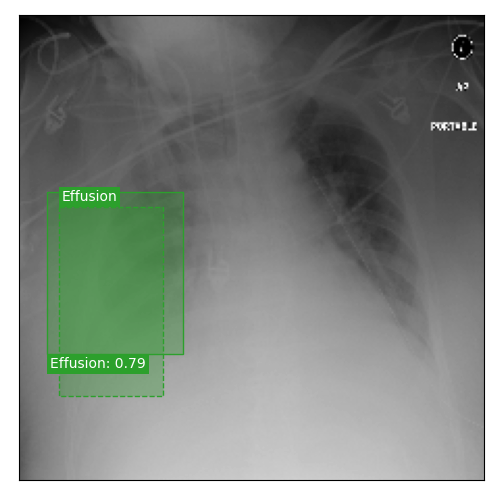} \\
    \end{tabular}
    \caption{Qualitative results of Loc-ADPD, with predicted (solid) and target (dashed) boxes. Cardiomegaly (red) is detected almost perfectly, as it is always exactly localized at one anatomical region. Other pathologies like atelectasis (blue), effusion (green), or pneumonia (cyan) are detected but often with non-perfect overlapping boxes. Detection also works well for predicting several overlapping pathologies (second from left).}
    \label{fig:qualitative}
\end{figure}
As shown in Fig. \ref{fig:qualitative} Loc-ADPD detects cardiomegaly almost perfectly, as it is always exactly localized at one anatomical region. Other pathologies are detected but often with too large or too small boxes as they only cover parts of anatomical regions or stretch over several of them, which cannot be completely corrected using WBF. Detection also works well for predicting several overlapping pathologies. For qualitative comparisons between Loc-ADPD and MIL-ADPD, we refer to Appendix \ref{app:qual}.

\section{Discussion and Conclusion}
\paragraph{Limitations}
While our proposed ADPD model outperforms all competing methods, it is still subject to limitations. First, due to the dependence on region proxies, they do not perform well on pathologies that only cover a small part of a region, as highlighted by the incapability of the models to detect nodules.
Additionally, while not requiring pathology bounding boxes, our models still require supervision in the form of anatomical region bounding boxes, and Loc-ADPD requires anatomy-level labels. However, anatomical bounding boxes are easier to annotate and predict than pathology bounding boxes, and the used anatomy-level labels were extracted automatically from radiology reports \cite{wu2021chest}. 

\paragraph{Conclusion}
We proposed a novel approach tackling pathology detection on chest X-rays using anatomical region bounding boxes. We studied two training approaches, using anatomy-level pathology labels and using image-level labels with MIL.
Our experiments demonstrate that using anatomical regions as proxies improves results compared weakly supervised methods and supervised training on little data, thus providing a promising direction for future research.

%
\bibliographystyle{splncs04}
\bibliography{ms.bib}

\begin{thebibliography}{10}
\providecommand{\url}[1]{\texttt{#1}}
\providecommand{\urlprefix}{URL }
\providecommand{\doi}[1]{https://doi.org/#1}

\bibitem{anaxnet}
Agu, N.N., Wu, J.T., Chao, H., Lourentzou, I., Sharma, A., Moradi, M., Yan, P.,
  Hendler, J.: Anaxnet: Anatomy aware multi-label finding classification in
  chest x-ray. In: de~Bruijne, M., Cattin, P.C., Cotin, S., Padoy, N., Speidel,
  S., Zheng, Y., Essert, C. (eds.) MICCAI. pp. 804--813 (2021).
  \doi{10.1007/978-3-030-87240-3_77}

\bibitem{DETR}
Carion, N., Massa, F., Synnaeve, G., Usunier, N., Kirillov, A., Zagoruyko, S.:
  End-to-end object detection with transformers. In: ECCV. p. 213–229 (2020).
  \doi{10.1007/978-3-030-58452-8_13}

\bibitem{weldon}
Durand, T., Thome, N., Cord, M.: Weldon: Weakly supervised learning of deep
  convolutional neural networks. In: CVPR. pp. 4743--4752 (2016).
  \doi{10.1109/CVPR.2016.513}

\bibitem{PhysioNet}
Goldberger, A.L., Amaral, L.A.N., Glass, L., Hausdorff, J.M., Ivanov, P.C.,
  Mark, R.G., Mietus, J.E., Moody, G.B., Peng, C.K., Stanley, H.E.:
  {PhysioBank, PhysioToolkit, and PhysioNet}: Components of a new research
  resource for complex physiologic signals. Circulation  \textbf{101}(23),
  e215--e220 (2000)

\bibitem{densenet}
Huang, G., Liu, Z., Van Der~Maaten, L., Weinberger, K.Q.: Densely connected
  convolutional networks. In: CVPR. pp. 2261--2269 (2017).
  \doi{10.1109/CVPR.2017.243}

\bibitem{stl}
Hwang, S., Kim, H.E.: Self-transfer learning for weakly supervised lesion
  localization. In: International conference on medical image computing and
  computer-assisted intervention. pp. 239--246. Springer (2016)

\bibitem{chexpert}
Irvin, J., Rajpurkar, P., Ko, M., Yu, Y., Ciurea{-}Ilcus, S., Chute, C.,
  Marklund, H., Haghgoo, B., Ball, R.L., Shpanskaya, K.S., Seekins, J., Mong,
  D.A., Halabi, S.S., Sandberg, J.K., Jones, R., Larson, D.B., Langlotz, C.P.,
  Patel, B.N., Lungren, M.P., Ng, A.Y.: Chexpert: A large chest radiograph
  dataset with uncertainty labels and expert comparison. In: AAAI. pp. 590--597
  (2019). \doi{10.1609/aaai.v33i01.3301590}

\bibitem{radgraph}
Jain, S., Agrawal, A., Saporta, A., Truong, S.Q.H., Duong, D.N., Bui, T.,
  Chambon, P.J., Zhang, Y., Lungren, M.P., Ng, A.Y., Langlotz, C.P., Rajpurkar,
  P.: Radgraph: Extracting clinical entities and relations from radiology
  reports. In: NeurIPS (2021)

\bibitem{johnson2019mimic}
Johnson, A.E.W., Pollard, T.J., Berkowitz, S.J., Greenbaum, N.R., Lungren,
  M.P., Deng, C.y., Mark, R.G., Horng, S.: Mimic-cxr, a de-identified publicly
  available database of chest radiographs with free-text reports. Scientific
  data  \textbf{6}(1), ~1--8 (2019)

\bibitem{johnson2019mimicphysio}
Johnson, A.E.W., Pollard, T.J., Berkowitz, S.J., Greenbaum, N.R., Lungren,
  M.P., Deng, C.y., Mark, R.G., Horng, S.: Mimic-cxr database (version 2.0.0).
  PhysioNet  (2019)

\bibitem{johnson2019mimicjpg}
Johnson, A.E.W., Pollard, T.J., Berkowitz, S.J., Greenbaum, N.R., Lungren,
  M.P., Deng, C.y., Mark, R.G., Horng, S.: Mimic-cxr-jpg, a large publicly
  available database of labeled chest radiographs. arXiv preprint
  arXiv:1901.07042  (2019)

\bibitem{adamw}
Loshchilov, I., Hutter, F.: Decoupled weight decay regularization. In: ICLR
  (2019)

\bibitem{lse}
Pinheiro, P.O., Collobert, R.: From image-level to pixel-level labeling with
  convolutional networks. In: CVPR. pp. 1713--1721 (2015).
  \doi{10.1109/CVPR.2015.7298780}

\bibitem{chexnet}
Rajpurkar, P., Irvin, J., Zhu, K., Yang, B., Mehta, H., Duan, T., Ding, D.,
  Bagul, A., Langlotz, C., Shpanskaya, K., et~al.: Chexnet: Radiologist-level
  pneumonia detection on chest x-rays with deep learning. arXiv preprint
  arXiv:1711.05225  (2017). \doi{10.48550/arXiv.1711.05225}

\bibitem{raoof2012interpretation}
Raoof, S., Feigin, D., Sung, A., Raoof, S., Irugulpati, L., Rosenow~III, E.C.:
  Interpretation of plain chest roentgenogram. Chest  \textbf{141}(2),
  545--558 (2012)

\bibitem{faster_rcnn}
Ren, S., He, K., Girshick, R., Sun, J.: Faster r-cnn: Towards real-time object
  detection with region proposal networks. NIPS  \textbf{28} (2015)

\bibitem{asl}
Ridnik, T., Ben-Baruch, E., Zamir, N., Noy, A., Friedman, I., Protter, M.,
  Zelnik-Manor, L.: Asymmetric loss for multi-label classification. In: ICCV.
  pp. 82--91 (2021). \doi{10.1109/ICCV48922.2021.00015}

\bibitem{gradcam}
Selvaraju, R.R., Cogswell, M., Das, A., Vedantam, R., Parikh, D., Batra, D.:
  Grad-cam: Visual explanations from deep networks via gradient-based
  localization. In: CVPR. pp. 618--626 (2017). \doi{10.1109/ICCV.2017.74}

\bibitem{wbf}
Solovyev, R., Wang, W., Gabruseva, T.: Weighted boxes fusion: Ensembling boxes
  from different object detection models. Image and Vision Computing
  \textbf{107},  104117 (2021).
  \doi{https://doi.org/10.1016/j.imavis.2021.104117}

\bibitem{cxr8}
Wang, X., Peng, Y., Lu, L., Lu, Z., Bagheri, M., Summers, R.M.: Chestx-ray8:
  Hospital-scale chest x-ray database and benchmarks on weakly-supervised
  classification and localization of common thorax diseases. In: CVPR. pp.
  2097--2106 (2017). \doi{10.1109/CVPR.2017.369}

\bibitem{wu2021chest}
Wu, J., Agu, N., Lourentzou, I., Sharma, A., Paguio, J.A., Yao, J.S., Dee,
  E.C., Kashyap, S., Giovannini, A., Celi, L.A., et~al.: Chest imagenome
  dataset for clinical reasoning. In: NIPS (2021)

\bibitem{wu2021chestphysio}
Wu, J.T., Agu, N.N., Lourentzou, I., Sharma, A., Paguio, J.A., Yao, J.S., Dee,
  E.C., Mitchell, W., Kashyap, S., Giovannini, A., et~al.: Chest imagenome
  dataset (version 1.0.0). PhysioNet  (2021)

\bibitem{wsup_thoracic}
Yan, C., Yao, J., Li, R., Xu, Z., Huang, J.: Weakly supervised deep learning
  for thoracic disease classification and localization on chest x-rays. In: ACM
  BCB. pp. 103--110 (2018)

\bibitem{agxnet}
Yu, K., Ghosh, S., Liu, Z., Deible, C., Batmanghelich, K.: Anatomy-guided
  weakly-supervised abnormality localization in chest x-rays. In: MICCAI. pp.
  658--668 (2022). \doi{10.1007/978-3-031-16443-9_63}

\bibitem{cam}
Zhou, B., Khosla, A., Lapedriza, A., Oliva, A., Torralba, A.: Learning deep
  features for discriminative localization. In: CVPR. pp. 2921--2929 (2016).
  \doi{10.1109/CVPR.2016.319}

\end{thebibliography}
\newpage
\appendix

\section{Detailed Results}\label{app:detailed_results}
\begin{table*}[h!]
    \centering
    \caption{Results per pathology of our models Loc-ADPD and MIL-ADPD as well as three relevant baselines, measured using the mAP metric.}
    \small
    \setlength{\tabcolsep}{3pt}
    \begin{tabular}{lccccc}
    \toprule
    Pathology & Loc-ADPD & MIL-ADPD & CheXNet & AGXNet & Faster R-CNN \\
    \midrule
    Atelectasis  & \textbf{2.27} & 1.15 & 1.96 & 1.28 & 0.00\\
    Cardiomegaly & \textbf{54.92} & 49.70 & 31.73 & 32.17 & 46.69 \\
    Effusion & \textbf{9.58} & 2.13 & 5.14 & 1.48 & 0.97 \\
    Infiltration & \textbf{5.19} & 1.43 & 3.00 & 3.61 & 0.93 \\
    Mass & \textbf{6.03} & 5.79 & 1.56 & 0.85 & 3.48 \\
    Nodule & 0.00 & 0.02 & 0.01 & 0.00 & \textbf{0.25} \\
    Pneumonia & \textbf{6.23} & 1.07 & 2.09 & 2.60 & 4.68 \\
    Pneumothorax & \textbf{2.92} & 1.42 & 0.96 & 0.42 & 1.93 \\
    \bottomrule
    \end{tabular}
\end{table*}
\FloatBarrier
\newpage 

\section{Qualitative Results and Failure Cases}\label{app:qual}
\begin{figure}[h!]
\small
\setlength{\tabcolsep}{0.2pt}
    \centering
    \begin{tabular}{@{}cc@{\hskip 0.5pt}cc@{}}
    Loc-ADPD & MIL-ADPD & Loc-ADPD & MIL-ADPD \\
    \addlinespace[2ex]
    \includegraphics[height=.24\textwidth]{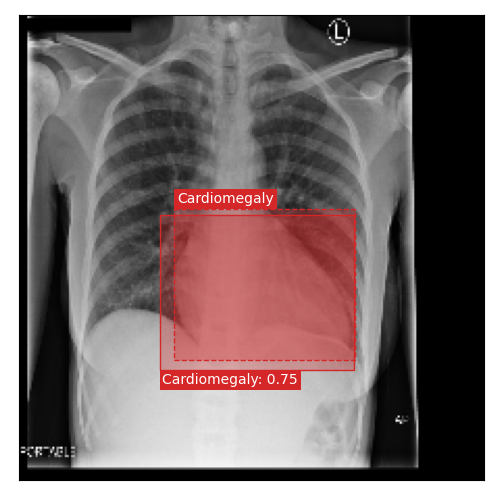} & 
    \includegraphics[height=.24\textwidth]{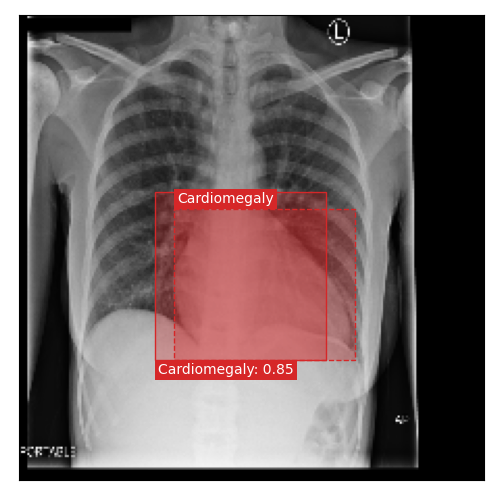} &
    \includegraphics[height=.24\textwidth]{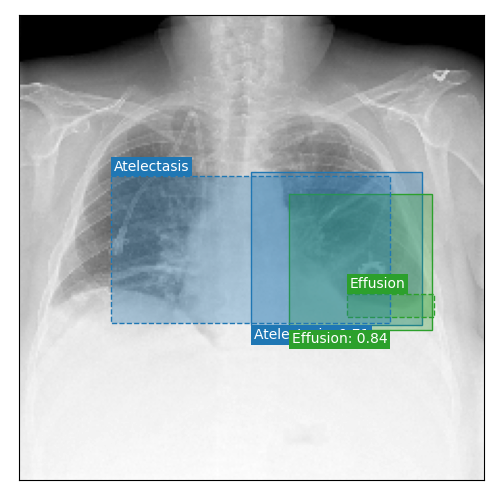} & 
    \includegraphics[height=.24\textwidth]{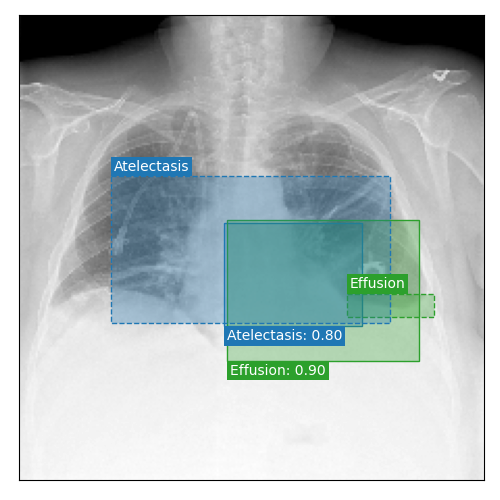} \\
    \cmidrule(lr){1-2}\cmidrule(lr){3-4}
    \multicolumn{2}{c}{(a)} & \multicolumn{2}{c}{(b)} \\
    \addlinespace[2ex]
    \includegraphics[height=.24\textwidth]{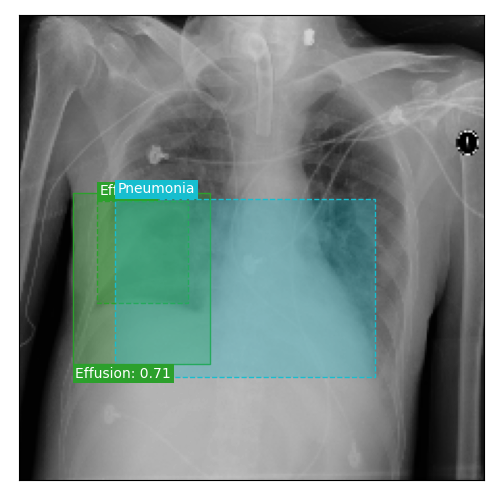}  &
    \includegraphics[height=.24\textwidth]{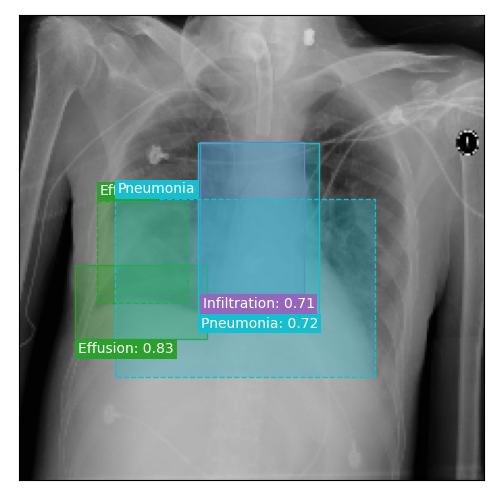}  &
    \includegraphics[height=.24\textwidth]{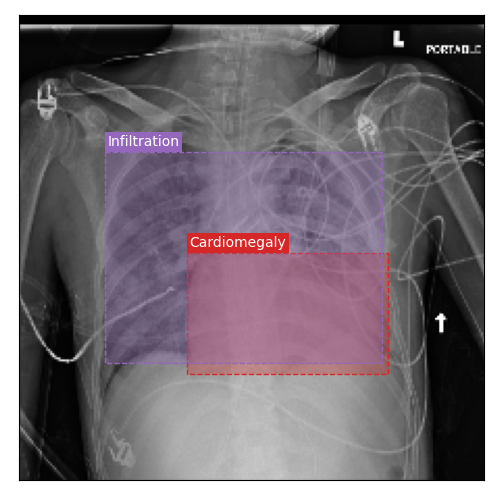}  &
    \includegraphics[height=.24\textwidth]{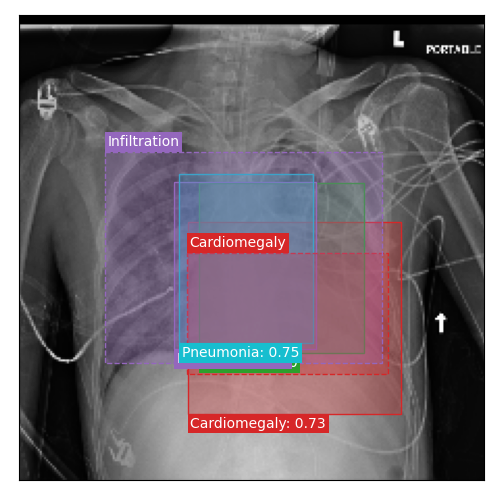}  \\
    \cmidrule(lr){1-2}\cmidrule(lr){3-4}
    \multicolumn{2}{c}{(c)} & \multicolumn{2}{c}{(d)} \\
    \addlinespace[2ex]
    \includegraphics[height=.24\textwidth]{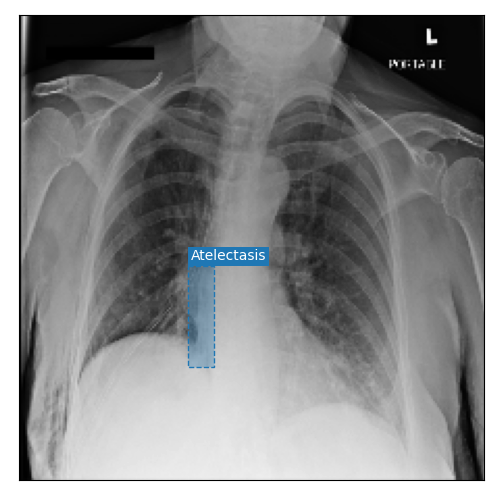}  &
    \includegraphics[height=.24\textwidth]{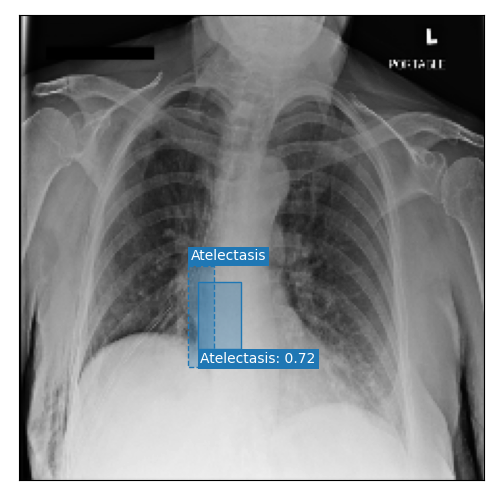}  &
    \includegraphics[height=.24\textwidth]{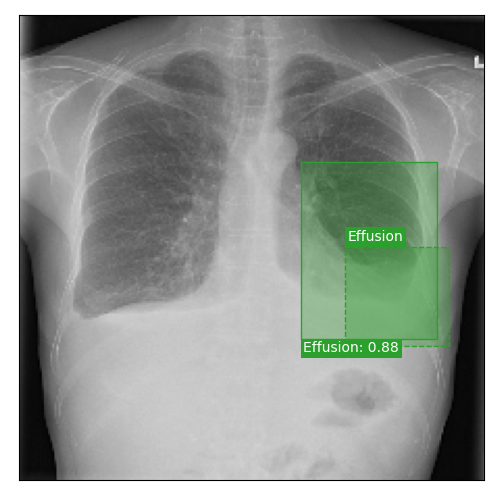}  &
    \includegraphics[height=.24\textwidth]{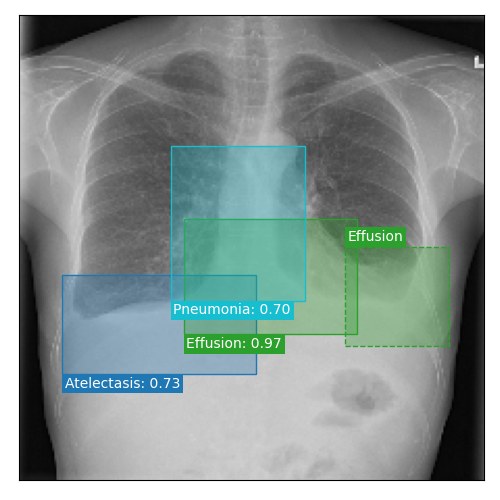}  \\
    \cmidrule(lr){1-2}\cmidrule(lr){3-4}
    \multicolumn{2}{c}{(e)} & \multicolumn{2}{c}{(f)} \\
    \end{tabular}
    \caption{Qualitative results and failure cases for the exemplary pathologies cardiomegaly (red), atelectasis (blue), effusion (green), pneumonia (cyan), and infiltration (purple).
    Loc-ADPD sometimes misses target boxes but is more accurate in general, while MIL-ADPD tends to overprediction.}
    \label{fig:qualitative2}
\end{figure}

\FloatBarrier
\newpage

\section{Class Mapping}\label{app:mapping}
\begin{table*}[h!]
    \centering
    \caption{Mapping between evaluation classes (bold) and training classes (included below the corresponding evaluation classes). Combined classes (+) represent averaging probabilities, while max represents taking the maximum of the related class probabilities. We show the number of training samples as well as the mAP results of our models. Used mappings are highlighted in grey, the best results per model are underlined.}
    \scriptsize
    \setlength{\tabcolsep}{3pt}
    \begin{tabular}{lccc}
    \toprule
    \textbf{Class (Pathology) Mapping} & \# Train & Loc-ADPD & MIL-ADPD  \\
    \midrule
    \textbf{Atelectasis} \\
    \rowcolor{Gray}Atelectasis & 79\,102 & \underline{2.27} & \underline{1.15} \\
    \midrule
    \textbf{Cardiomegaly} \\
    \rowcolor{Gray}Enlarged cardiac silhouette & 57\,565 & \underline{54.92} & \underline{49.70} \\
    \midrule
    \textbf{Effusion} \\
    \rowcolor{Gray}Pleural Effusion & 3\,489 & \underline{9.58} & \underline{2.13} \\
    \midrule
    \textbf{Infiltration} & & \\
    Infiltration & 2\,973 & 0.00 & 0.46 \\
    Lung Opacity & 154\,825 & 4.56 & 1.38 \\
    \rowcolor{Gray}Infiltration + Lung Opacity & 154\,825 & \underline{5.19} & \underline{1.43} \\
    Infiltration + Lung Opacity (max) & 154\,825 & 4.56 & 1.38 \\
    \midrule
    \textbf{Mass}  \\
    Mass/Nodule & 7\,804 & 6.17 & 5.11 \\
    Multiple masses/nodules & 3734 & 4.38 & 5.43 \\
    Lung opacity & 154\,825 & 0.86 & 0.77 \\
    Lung lesion & 13\,481 & 5.59 & 5.49 \\
    \rowcolor{Gray}Mass/Nodule + Multiple masses/nodules & 10\,838 & 6.03 & \underline{5.79} \\
    Mass/Nodule + Multiple masses/nodules (max) & 10\,838 & \underline{6.19} & 5.11 \\
    \makecell[l]{Mass/Nodule  + Multiple masses/nodules \\ + Lung opacity} & 154\,825 & 5.77 & 4.56\\
    \makecell[l]{Mass/Nodule + Multiple masses/nodules \\+ Lung lesion} & 13\,481 & 5.61 & 5.79 \\
    \midrule
    \textbf{Nodule} & \\
    \rowcolor{Gray}Mass/Nodule & 7\,804 & 0.00 & \underline{0.02} \\
    Multiple masses/nodules & 3\,734 & 0.00 & 0.01 \\
    Lung opacity & 154\,825 & 0.00 & 0.00 \\
    Lung lesion & 13\,481 & 0.00 & 0.02 \\
    Mass/Nodule + Multiple masses/nodules & 10\,838 & 0.00 & 0.02 \\
    \makecell[l]{Mass/Nodule + Multiple masses/nodules \\+ Lung opacity} & 154\,825 & 0.00 & 0.02 \\
    \makecell[l]{Mass/Nodule + Multiple masses/nodules \\+ Lung lesion} & 13\,481 & 0.00 & 0.02 \\
    \midrule
    \textbf{Pneumonia}  \\
    \rowcolor{Gray}Pneumonia & 154\,825 & \underline{6.23} & \underline{1.07} \\
    \midrule
    \textbf{Pneumothorax} \\
    \rowcolor{Gray}Pneumothorax & 10\,435 & \underline{2.92} & \underline{1.42} \\
    \bottomrule
    \end{tabular}
\end{table*}

\end{document}